\documentclass{article}
\usepackage[final]{colm2026_conference}

\usepackage{latexsym}
\usepackage[T1]{fontenc}
\usepackage[utf8]{inputenc}
\usepackage{microtype}
\usepackage{inconsolata}
\usepackage{graphicx}
\usepackage{booktabs}
\usepackage{multirow}
\usepackage{amsmath}
\usepackage{amssymb}
\usepackage{algorithm}
\usepackage{algpseudocode}
\usepackage{xcolor}
\usepackage[hidelinks]{hyperref}
\usepackage{url}

\makeatletter
\newcommand{\blfootnote}[1]{%
  \begingroup
    \renewcommand{\thefootnote}{}%
    \renewcommand{\@makefntext}[1]{\noindent##1}%
    \footnote{#1}%
    \addtocounter{footnote}{-1}%
  \endgroup
}
\makeatother
\title{The Lifecycle of LLM-as-a-Judge for Large-Scale \\
  Recommendation Explanations}

\author{Emma Yanyang Kong \quad JJ Tan \quad
Ishan Gupta\textsuperscript{\dag} \quad David Fagnan \quad Lars Olds \\ 
\bfseries Claire Campbell \quad Ratna Kavuri \quad Veli Balin \quad Rohan Gosain \quad Louis Garcia \\
\bfseries Minsu Jang \\[4pt]
{\normalfont Netflix, Los Gatos, California, USA} \\[3pt]
{\normalfont\small\textsuperscript{\dag}Work done while at Netflix.}}

\begin{document}
\maketitle

% Override the template's default "Published as a conference paper" banner:
% both venues are non-archival COLM 2026 workshops.
\lhead{Accepted at the Lifelong Agents and AIMS workshops at COLM 2026}

\blfootnote{\scriptsize Accepted at the 2nd Workshop on Lifelong Agents:
Learning, Aligning, and Evolving and at the AIMS Workshop (\emph{AI
Measurement Science: Toward Rigorous AI Evaluation}), both co-located
with COLM 2026.}

% ===============================================================
\begin{abstract}
LLM-as-a-Judge, which leverages a large language model to evaluate
natural language generated by another AI application or model, has
become a standard, scalable approach for accelerating and extending
costly human evaluation. Yet most work treats a judge as a static
artifact, evaluating it once at construction or against a fixed
benchmark. We argue instead that an LLM judge operating in a deployed
system is better understood as having a \emph{lifecycle}. It must be
built, trained, deployed, and continuously maintained as the surrounding
data evolves, and each phase poses distinct technical and operational
challenges.

We present such a lifecycle for the LLM judges that evaluate
recommendation explanations at Netflix. Everything we report comes out of
a series of controlled online member-facing experiments, in which our pipeline
generated and the judges assessed hundreds of thousands of distinct
show-level explanations per week across a changing catalog. Our framework has
four phases. \textbf{(I)~Birth} defines the evaluation criteria and
builds curated benchmark datasets with human labels and rationales.
\textbf{(II)~Training} refines the judges' rubrics via
\emph{Reasoning-Aligned Rubric Tuning} (RART), which uses a meta-judge
over reasoning output as the learning signal. \textbf{(III)~Deployment}
puts one judge in two online roles, quality gating and reflective
generation. \textbf{(IV)~Monitoring} runs a continuous Human-in-the-Loop (HITL)
alignment process that detects drift and triggers re-tuning behind a
human review gate. We report results from a five-week
online A/B test over tens of millions of members on the Netflix mobile
app. Against a no-explanation control, judge-aligned explanations shifted
member viewing toward novel content (previously unwatched) and increased
successful browse-to-play sessions, with no quality-related escalations.
\end{abstract}

% ===============================================================
\section{Introduction}
\label{sec:intro}

A recommendation explanation, a short natural-language message
accompanying a recommended item, shapes how users perceive and
trust a recommender system. In our work, an explanation
is human-readable evidence for \emph{why} a title was recommended based on
what the member has previously watched. We focus on
\emph{similarity-based} explanations, which connect a recommended
title to one reference title the member has interacted with via shared
attributes such as genre and tone, e.g.,
``A funny, heartfelt holiday romance about love and new
beginnings, much like \textit{My Secret Santa}.''
These explanations aim to drive \emph{content discovery}
and build \emph{trust} for our members, which is also why quality matters. A misleading or poorly grounded
explanation erodes the very trust the feature is meant to build.

Evaluating quality at scale is hard. In our experiments the pipeline
produced hundreds of thousands of distinct item-level explanations per
week, any of which could reach members, and each must be accurate,
item-specific, and free of sensitive or offensive content. Human
evaluation, the gold standard, is not feasible at this volume.
LLM-as-a-Judge~\citep{zheng2023judging,liu2023geval,kim2024prometheus}
offers a scalable proxy, but a judge is not a one-shot artifact. It must
be initialized against trusted human labels, tuned to align with humans,
deployed where it acts on live traffic, and maintained as the
upstream data drifts. To our knowledge, no prior work characterizes all
four phases as a single, instrumented lifecycle.

We therefore treat the judge not as a fixed evaluator but as a
\emph{lifelong agent}. It is a persistent component that must continually
\emph{learn} from accumulating human feedback and remain \emph{aligned}
with human judgment for the right reasons rather than by coincidence. It
must also \emph{evolve} through standing alignment audits and automated
re-tuning, because the catalog, recommendation algorithms, and user
population all shift.

In this paper, we present a case study of an LLM judge for
recommendation explanations at Netflix,
framed as a lifecycle (Figure~\ref{fig:lifecycle}) with four phases:
\textbf{(I)~Birth} (\S\ref{sec:birth}),
\textbf{(II)~Training} (\S\ref{sec:training}),
\textbf{(III)~Deployment} (\S\ref{sec:deployment}), and
\textbf{(IV)~Monitoring} (\S\ref{sec:stewardship}). Our contributions are:
\begin{enumerate}\itemsep2pt
  \item A lifecycle view of LLM judges in a large-scale recommender
    system, covering all four phases end to end.
  \item Reasoning-Aligned Rubric Tuning (RART), a rubric-refinement
    method that iteratively updates the judge's rubric using a meta-judge
    over the judge's reasoning, with an ablation isolating what reasoning
    alignment contributes.
  \item An operational design for serving, pairing a two-role
    deployment (quality gate and critic in a self-reflection revision
    loop) with continuous drift monitoring that triggers re-tuning behind
    a human review gate.
  \item Practical takeaways for shipping LLM judges at scale.
\end{enumerate}

% ===============================================================
\section{Related Work}
\label{sec:related}

\paragraph{Explainable recommendation.} Explaining \emph{why} an item was
recommended is a long-standing concern, from surfacing
collaborative-filtering neighborhoods~\citep{herlocker2000explaining} to
surveys organizing explanation aims and their
evaluation~\citep{tintarev2007survey,zhang2020explainable} around goals
such as transparency, trust, and persuasiveness. Our similarity-based
explanations sit in the transparency-and-trust branch of that taxonomy.
What differs here is not the explanation style but the evaluation problem
it creates. Free-form LLM text at catalog scale forfeits the
template-level correctness earlier systems guaranteed by construction, so
quality must itself be measured rather than assumed.

\paragraph{LLM-as-a-Judge.} Using an LLM to score another model's output
is now a standard substitute for costly human evaluation in open-ended
generation~\citep{zheng2023judging,liu2023geval,zhu2023judgelm,kim2024prometheus}.
Such judges carry systematic biases, including sensitivity to answer
position and verbosity~\citep{wang2024fair} and a preference for their
own generations~\citep{panickssery2024selfpref}, which is one reason we
anchored ours to human labels and rationales at every phase. Public
benchmarks such as JudgeBench~\citep{tan2024judgebench} and
RewardBench~\citep{lambert2024rewardbench} score judges on
general-purpose correctness, safety, and instruction-following, but they
are static and domain-general, and none treats the judge as something
with a lifecycle to be re-validated after deployment. Our Phase~I
benchmark addresses that gap with a domain-specific,
rationale-annotated dataset refreshed from live traffic.

\paragraph{Self-refinement and reflection.} Using a model's own critique
to improve its output without gradient updates has been explored for the
generator (Self-Refine~\citep{madaan2023selfrefine},
Reflexion~\citep{shinn2023reflexion}) and, closer to our setting, for the
judge. Meta-Rewarding~\citep{wu2024metarewarding} adds a meta-judge
scoring the judge's own reasoning. Our reasoning meta-judge $\mathcal{M}$
(\S\ref{sec:reasoning-alignment}) builds directly on this idea, with two differences
that matter. First, it is grounded in human
rationales rather than unsupervised self-consistency. Second, it deliberately targets fail
examples, because in Phase~III the judge's rejection reason becomes the
generator's revision instruction, so a right-verdict but wrong-reason
case propagates a misleading signal downstream.

\paragraph{Prompt and rubric optimization.} Textual analogues of
gradient-based optimization revise a prompt or rubric from feedback
rather than backpropagating through weights. They span evolutionary prompt search
(EvoPrompt~\citep{guo2024evoprompt},
Promptbreeder~\citep{fernando2023promptbreeder}), textual gradients
(TextGrad~\citep{yuksekgonul2024textgrad}, ACE~\citep{zhang2025ace}),
reflective evolution over a Pareto pool (GEPA~\citep{agrawal2026gepa}, on
DSPy~\citep{khattab2024dspy}), and rubric synthesis from chosen/rejected
pairs~\citep{liu2025openrubrics}. RART (\S\ref{sec:art}) belongs to this
family, closest to GEPA but a greedy, single-objective special case
(\S\ref{sec:future}). Its textual gradient is also computed over
\emph{rationale} mismatches, so that it targets reasoning quality and not
just label accuracy. 
None of this work evaluates a judge continuously
after deployment or ties re-tuning to a monitored drift signal. Our
contribution is orthogonal and compatible. We show what changes when a
judge is one node in a closed-loop live system rather than scored once
against a fixed benchmark.

% ===============================================================
\section{System Overview}
\label{sec:overview}

\begin{figure}[t]
\centering
\includegraphics[width=0.92\textwidth]{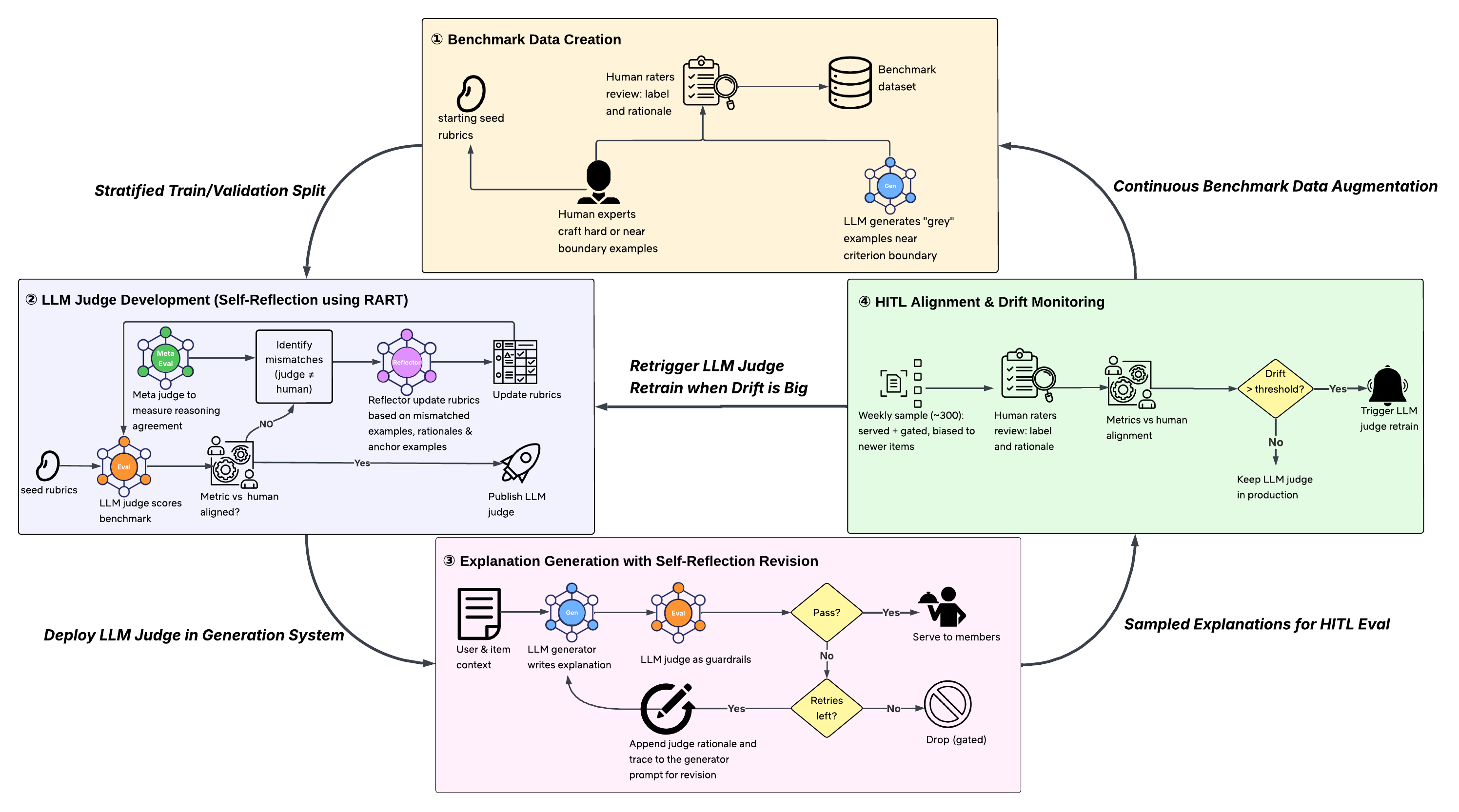}
\caption{The four-phase lifecycle of an LLM judge for recommendation
explanations. \textbf{(I)~Birth} (\S\ref{sec:birth}) builds a benchmark
of human-labeled, rationale-annotated examples. \textbf{(II)~Training}
(\S\ref{sec:training}) has a reflector LLM tune each criterion's rubric
from judge--human label and \emph{reasoning} mismatches.
\textbf{(III)~Deployment} (\S\ref{sec:deployment}) gates generated
explanations and drives self-reflective revision with bounded retries.
\textbf{(IV)~Monitoring} (\S\ref{sec:stewardship}) uses a weekly
human-rated sample to detect drift and augment the benchmark,
re-triggering (II) past threshold.}
\label{fig:lifecycle}
\end{figure}

Figure~\ref{fig:lifecycle} summarizes the four phases and the artifacts
flowing between them. LLM judges play three roles, the optimization
target of Phase~II, the feedback provider and guardrail decision-maker of
Phase~III, and the monitored object of Phase~IV. Human labels and
rationales are collected in Phase~I and continuously augmented at
$\sim$300/week in Phase~IV. That phase closes two loops, a fast
drift-detection loop that re-triggers Phase~II re-training when
judge--human agreement decays, and a slower benchmark-augmentation loop
that keeps the Phase~I dataset representative of the live catalog. Over the five-week A/B test of
\S\ref{sec:abtest}, the system evaluated hundreds of thousands of
explanations per week, passing over 75\% with a retry budget of 3.

% ===============================================================
\section{Phase I: Birth, Establishing Ground Truth}
\label{sec:birth}

The Birth phase is the most human-intensive of the four. Our internal
writing experts define the criteria, author the labeling
guidelines for human raters, hand-craft adversarial examples, and anchor
human rating. The guidelines double as the seed rubrics for the LLM
judges optimized in Phase~II. Raters are trained and calibrated by our
Human Data Operations team, whose protocol is internal and not detailed
here. Every fail label carries a free-text rationale against the
guideline.

Public LLM-judge benchmarks~\citep{tan2024judgebench,lambert2024rewardbench}
target general-purpose evaluation and miss the constraints of
recommendation explanations, which are item-specific,
context-dependent, and short. We therefore build a domain-specific
benchmark of similarity-based explanations, each conditioned on a target
item and one or two reference items the user has interacted with in the
past. The criteria split into \emph{must-have} pass/fail conditions that
every explanation must satisfy before being shown, and softer stylistic
and recommendation-relevance criteria (the specific criteria are
withheld for confidentiality). For each criterion, the labeling
guidelines, pass/fail conditions, and boundary examples are the sources
of truth for both the LLM judges and the human raters.

The benchmark draws on three complementary sources:
(i) expert-crafted examples, each with a human rater's pass/fail label,
its known failure mode, and a rationale for failure, covering cases the
other two sources cannot reliably produce;
(ii) human-rated, LLM-synthesized examples near the criterion boundary,
capturing difficult cases that naturalistic sampling rarely surfaces; and
(iii) sampled explanations from our explanation-generation system,
collected before it was fully integrated into the serving pipeline.

At the start of the online experiments, the benchmark held roughly 900 human-labeled
explanations, split close to evenly between the two labels with a slight
majority of \textsc{fail} ($\sim$54\%). We deliberately hold the classes
near balance rather than matching the prevalence in generated samples,
where failures are far rarer. Specificity (Eq.~\ref{eq:spec}),
$\mathrm{RA}_{\mathrm{neg}}$ (Eq.~\ref{eq:raneg}), and the training data
of Algorithm~\ref{alg:art} all depend on human-failed examples. As reported for Phase~II, the alignment
metrics of \S\ref{sec:metrics} therefore measure agreement on this
class-balanced, difficulty-enriched benchmark, not defect rates
on live traffic. 

The dataset is not fixed. The Phase~IV Human-in-the-Loop (HITL) pipeline
(\S\ref{sec:stewardship}) appends $\sim$300 freshly rated explanations
each week during tests, tracking distribution shift in the live catalog.

% ===============================================================
\section{Phase II: Training, Aligning the Judge to Humans}
\label{sec:training}

We tune one judge per \emph{must-have} criterion (\S\ref{sec:birth}),
three in total, using self-reflection to iteratively refine that
criterion's rubric against human labels and their failure rationales. We call this procedure
\textbf{Reasoning-Aligned Rubric Tuning} (RART). One key component of RART enforces 
\emph{reasoning alignment} between the LLM judge and human raters.

\subsection{Reasoning-Aligned Rubric Tuning (RART)}
\label{sec:art}

Each LLM judge is instantiated from a small, task-invariant prompt
template. A fixed system message describes the evaluation task and
carries a templated \texttt{<criterion>} slot holding the rubric being
tuned; a user message supplies the explanation text and the target- and
reference-item metadata. The judge emits a JSON object
\{\texttt{label}, \texttt{reason}\} per explanation, so tuning reduces to
refining the criterion rubrics while the surrounding prompt is held
constant.

Given a rubric $R_t$ at iteration $t$, RART iterates the loop in
Algorithm~\ref{alg:art}. It scores the data with the current judge
$J(R_t)$, checks validation metrics for early stopping, and otherwise
asks a reflector LLM~$\mathcal{R}$ to propose a refined rubric from the
judge's errors. Crucially, an ``error'' is not only a wrong label. Before
reflecting, a \emph{rationale meta-judge}~$\mathcal{M}$ takes the
examples that \emph{both} the judge and the human labeled \emph{fail} and
compares the judge's free-text reason against the human rationale
(\S\ref{sec:reasoning-alignment}). The reflector therefore targets two
error types, examples the judge mislabels and agreed-fail examples where
it reaches the right verdict for the wrong reason. We write
$\ell_J(x),\ell_H(x)$ for the judge and human labels on example $x$, and
$r_J(x),r_H(x)$ for their respective reasons.

\begin{algorithm}[t]
\caption{Reasoning-Aligned Rubric Tuning}
\label{alg:art}
\scriptsize
\begin{algorithmic}[1]
\Require initial rubric $R_0$; train/val/test sets
  $\mathcal{D}_{\text{tr}},\mathcal{D}_{\text{val}},\mathcal{D}_{\text{te}}$;
  max iterations $N$
\Statex \textbf{Judges:} rubric-conditioned judge $J(R)\!:\!x\mapsto(\ell_J,r_J)$ (label, rationale);
  rationale meta-judge $\mathcal{M}(r_J,r_H)\!\to\!\{\textsc{agree},\textsc{mismatch}\}$;
  reflector $\mathcal{R}(R,\textit{focus})\!\to\!R'$ (revised rubric)
\Ensure tuned rubric $R^\star$
\State $R^\star \gets R_0$;\quad $s^\star \gets -\infty$
\For{$t = 0$ \textbf{to} $N-1$}
  \State Score $\mathcal{D}_{\text{tr}},\mathcal{D}_{\text{val}}$ with $J(R_t)$
         \Comment{each $x$ gets $(\ell_J,r_J)$}
  \State $s \gets$ weighted alignment metrics on $\mathcal{D}_{\text{val}}$ (\S\ref{sec:metrics})
  \If{$s > s^\star$}
    \State $R^\star \gets R_t$;\quad $s^\star \gets s$
  \EndIf
  \If{all metrics clear the per-criterion bound}
    \State \textbf{break}
  \EndIf
  \State $\mathcal{X} \gets \{x : \ell_J(x) \neq \ell_H(x)\}$
         \Comment{label mismatches}
  \State $\mathcal{N} \gets \{x : \ell_J(x) = \ell_H(x) = \textsc{fail}\}$
         \Comment{agreed-fail (negative) examples}
  \State $\mathcal{N}_{\!\times} \gets \{x \in \mathcal{N} : \mathcal{M}(r_J(x),r_H(x))=\textsc{mismatch}\}$
         \Comment{agreed-fail but wrong reason}
  \State $\textit{focus} \gets \mathcal{X} \cup \mathcal{N}_{\!\times}$
         \Comment{label mismatch, or agreed-fail but wrong reason}
  \State $R_{t+1} \gets \mathcal{R}(R_t,\ \textit{focus})$
         \Comment{sample candidate rubric}
\EndFor
\State \Return $R^\star$
\end{algorithmic}
\end{algorithm}

\subsection{Alignment Metrics}
\label{sec:metrics}

We use three metrics aggregated over the benchmark dataset. Write
$\ell_J(x),\ell_H(x)\in\{\textsc{pass},\textsc{fail}\}$ for the judge
and human labels on example $x$.

\textbf{Specificity} (\emph{fail-recall}) is the judge's rate of correctly
rejecting bad explanations. It is our most critical metric, since bad
explanations that slip past the gate directly erode member trust:
\begin{equation}
  \mathrm{Spec} \;=\;
  \frac{|\{x : \ell_J(x)=\textsc{fail},\,\ell_H(x)=\textsc{fail}\}|}
       {|\{x : \ell_H(x)=\textsc{fail}\}|}.
  \label{eq:spec}
\end{equation}

\textbf{Recall} (\emph{pass-recall}) is the judge's rate of correctly passing
good explanations. High recall is also needed to keep explanation
coverage high and revision cost low:
\begin{equation}
  \mathrm{Rec} \;=\;
  \frac{|\{x : \ell_J(x)=\textsc{pass},\,\ell_H(x)=\textsc{pass}\}|}
       {|\{x : \ell_H(x)=\textsc{pass}\}|}.
  \label{eq:rec}
\end{equation}

\textbf{Reasoning agreement rate} ($\mathrm{RA}_{\mathrm{neg}}$) is the
fraction of human-failed examples
$\mathcal{F}=\{x:\ell_H(x)=\textsc{fail}\}$ on which the judge and the human not
only both reject the explanation but also agree on \emph{why} it fails.
The numerator is restricted to the \emph{agreed-fail} set
$\mathcal{N}=\{x:\ell_J(x)=\ell_H(x)=\textsc{fail}\}\subseteq\mathcal{F}$,
since reasoning can only agree where both labels are \textsc{fail}, while
the denominator is all of $\mathcal{F}$. Agreement is decided by the
rationale meta-judge $\mathcal{M}$:
\begin{equation}
  \mathrm{RA}_{\mathrm{neg}} \;=\;
  \frac{|\{x \in \mathcal{N} :
        \mathcal{M}(r_J(x),r_H(x))=\textsc{agree}\}|}{|\mathcal{F}|}.
  \label{eq:raneg}
\end{equation}

For rubric \emph{optimization} during RART we collapse these into a single
weighted score
\begin{equation}
  s \;=\; w_s\,\mathrm{Spec} + w_r\, \mathrm{Rec}
          + w_{ra}\,\mathrm{RA}_{\mathrm{neg}}.
  \label{eq:objective}
\end{equation}
We weight specificity above the other two terms, $w_s=3$ and
$w_r=w_{ra}=1$, because the two error types are not symmetric
online. A bad explanation that escapes gating is served to members,
whereas a good one that is rejected is only revised or dropped
(\S\ref{sec:guardrail}). The $\mathrm{RA}_{\mathrm{neg}}$ term
additionally favors rubrics whose rejections are backed by the right
reason rather than by coincidence.

\subsection{Reasoning-Aligned Self-Reflection}
\label{sec:reasoning-alignment}

\begin{figure}[t]
\centering
\includegraphics[width=\textwidth]{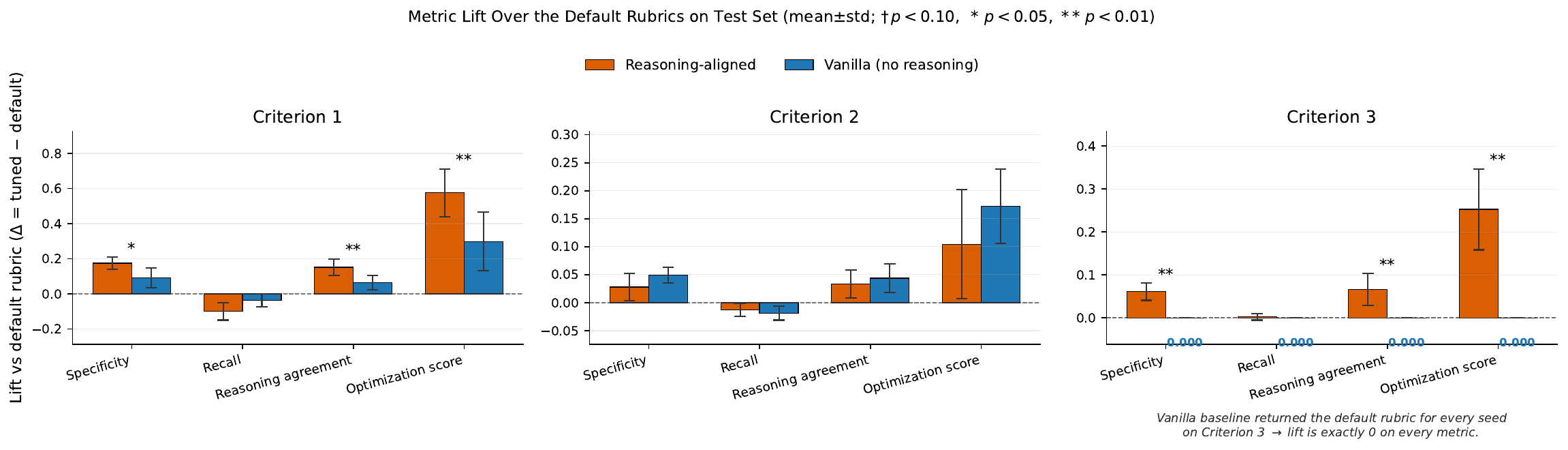}
\caption{Alignment-metric lift over the default rubric on the held-out
test set for RART (rationale-aware reflection) vs.\ vanilla (label-only
reflection), one panel per must-have criterion, $n{=}8$ seeds each reshuffling the train/validation/test
split; bars show mean, error bars $\pm$std. All quantities are
$\Delta = \text{tuned} - \text{default}$, so positive means improvement.
Significance on $\Delta$ (RART$-$vanilla) by two-sided sign test:
$^{\dag}$~$p{<}0.10$, $^{*}$~$p{<}0.05$, $^{**}$~$p{<}0.01$.}
\label{fig:rart-vs-art-lift}
\end{figure}

Treating every disagreement uniformly ignores that mismatches differ in
information content. The judge may agree on the label for the wrong
reason, or disagree on the label while its rationale exposes a genuine
rubric ambiguity. Building on the \emph{meta-judge}
idea~\citep{wu2024metarewarding}, we introduce a \emph{reasoning
meta-judge} $\mathcal{M}$ that compares the judge's reason against the
human rationale and returns \textsc{rationale\_agreement} or
\textsc{rationale\_mismatch}. We run $\mathcal{M}$ only on agreed-fail
examples, where a shared ``fail'' label can still hide divergent reasons.
The rationale mismatches it finds join the label mismatches in the
reflector's focus set, and agreements are left untouched so the rubric
preserves what the judge already gets right.

Restricting $\mathcal{M}$ to agreed-fail examples is also a deployment
requirement. In Phase~III (\S\ref{sec:deployment}) the same judge is the
critic in a self-reflection loop, where its rejection reason becomes the
generator's revision instruction, so a right-verdict but wrong-reason
rejection propagates a misleading signal. Unlike
Meta-Rewarding~\citep{wu2024metarewarding}, whose meta-judge is
unsupervised, ours is grounded in human rationales. The signal is a text
``gradient'' in the spirit of TextGrad~\citep{yuksekgonul2024textgrad},
ACE~\citep{zhang2025ace}, and GEPA~\citep{agrawal2026gepa}, but computed
over rationales rather than generations, and complementary to rubric
synthesis from chosen/rejected pairs~\citep{liu2025openrubrics}.

\paragraph{Validating the meta-judge.}
$\mathcal{M}$ supplies both our reasoning-agreement metric and the
training signal for RART, so we validate it against humans directly.
Trained raters independently labeled rationale agreement on a sample of
300 agreed-fail rationale pairs, and $\mathcal{M}$'s verdicts matched
their judgments $98.6\%$ of the time, supporting its use as both an
evaluation metric and a learning signal.

To isolate the contribution of the reasoning signal, we compare RART
against \emph{vanilla}, an identical loop whose reflector sees only label
mismatches, on each of the three must-have criteria
(Figure~\ref{fig:rart-vs-art-lift}; $8$ seeds, held-out test).
RART improves our most important metric, \emph{specificity}, more than
vanilla wherever the default rubric leaves headroom. On criterion~1 this
comes with a recall cost, which is
acceptable because a falsely rejected explanation re-enters the revision
loop (\S\ref{sec:guardrail}) and is regenerated, whereas a bad
explanation that reaches members cannot be recalled. On criterion~3,
label-only vanilla collapsed specificity and reasoning agreement in every
iteration. The best-checkpoint rule of Algorithm~\ref{alg:art} therefore
returned essentially the default rubric, leaving its lift at $\approx 0$,
whereas RART lifts both metrics and even improves recall slightly. On
criterion~2 the default rubric is already near ceiling and the two
methods are indistinguishable. Reasoning alignment therefore helps where there is room and does not
hurt where there is none, and across all three must-have criteria the headroom the
default rubric leaves is what governs how much RART can gain.

% ===============================================================
\section{Phase III: Deployment, Putting the Judge to Work}
\label{sec:deployment}

The tuned judge from \S\ref{sec:training} serves two online roles on
every generated explanation. As a \emph{gate} it rejects explanations
failing a configurable subset of criteria. As the \emph{critic} in a
self-reflective revision loop it returns flagged explanations to the
generator. Explanations are generated per recommended item rather than
per (user, item) pair; an online personalization model then selects the
best-fitting explanation for each pair. A single explanation can
therefore be shared across many users.

\subsection{Judge as Guardrail with Bounded-Retry Revision}
\label{sec:guardrail}

In the serving pipeline every explanation passes through a
\emph{generate $\rightarrow$ judge $\rightarrow$ revise} loop with a fixed
retry budget. The generator drafts an explanation conditioned on the style (which
captures the user cohort), target item, and reference item(s); the tuned judge scores it
against the criteria of \S\ref{sec:birth}, emitting a pass/fail
label and reason per criterion. If all criteria pass, the explanation is
\emph{served}; if any fails, the judge's reason is appended to the
generator prompt and the generator is re-invoked, up to $K$ retries, and
an explanation still failing after $K$ attempts is \emph{dropped}. We
write $k$ for the retry index and $K$ for the deployed budget. Dropping
rather than serving a flagged explanation is a deliberate asymmetry. A
bad explanation is a trust hazard (\S\ref{sec:metrics}), whereas a
missing one merely forgoes an opportunity.

\begin{figure}[t]
\centering
\includegraphics[width=0.5\textwidth]{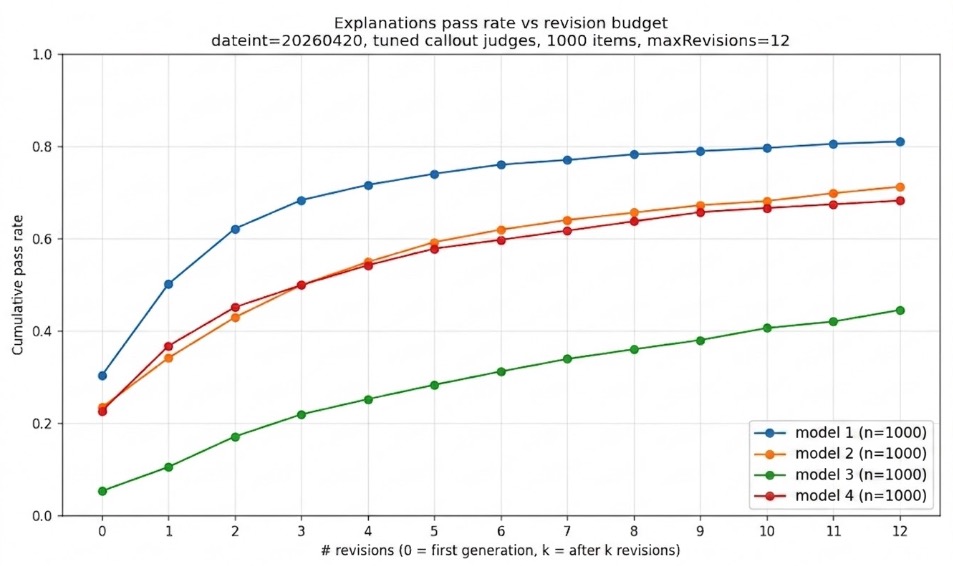}
\caption{Cumulative judge pass rate vs.\ revision budget $k$ on
$n{=}1000$ generated explanations across four generator models.
Gains are monotonic but flatten beyond $k{\approx}4$. The weakest
generator (model~3) stays well below the others. This implies judge-guided
revision amplifies a capable generator rather than substituting for one.}
\label{fig:pass-rate}
\end{figure}

\paragraph{Cumulative pass rate vs.\ revision budget.}
Figure~\ref{fig:pass-rate} reports cumulative judge pass rate versus
retry budget $k$ on a sample of $n{=}1000$ generated explanations across
four generator models. Pass rate rises monotonically in $k$ because the
judge's free-text \texttt{reason} steers the next draft, but returns
diminish fast. The three strong generators capture $\geq 80\%$ of their
achievable lift by $k{=}3$--$4$, which motivates a small fixed $K$.
Revision only partly rescues weak generators (model~3 stays below 50\%
even at $k{=}12$), so a sustained drop in the $k{=}0$ pass rate signals a
generator-side regression rather than judge drift. We deploy $K{=}3$ with
the strongest model, which captures most of the achievable lift while
bounding per-explanation cost; at this budget the end-to-end pipeline
runs at a few thousand US dollars per week in inference cost.

\subsection{Online A/B Evaluation}
\label{sec:abtest}

High offline judge--human alignment and low judged defect rates are
\emph{necessary} for explanations to benefit users, but not
\emph{sufficient}. They certify that served explanations meet our quality
bar, not that having explanations improves user experience. To test the
latter directly, we ran a large-scale A/B test on the mobile surface,
comparing the judge-aligned explanation pipeline
(\S\ref{sec:training}--\ref{sec:deployment}) against a no-explanation
control over tens of millions of members for five weeks. Treatment shifted
member viewing toward \emph{novel content}, i.e., titles the member had
not previously watched, by $+0.2\%$ relative ($p < 0.05$), consistent
with explanations helping members discover unfamiliar titles rather than
only reinforcing already-familiar viewing. We also observed a $+0.3\%$
relative increase in \emph{sessions with a successful play} ($p < 0.05$),
indicating that members more often found something to watch during a
browse session, which we read as reduced browse friction. These relative
lifts are small in absolute value, but at our scale they correspond to a
substantial aggregate effect, and movements of this magnitude are
considered meaningful for a feature-level intervention on this surface.
Judge-aligned explanations are therefore not just
\emph{quality-compliant} but also \emph{useful}. As further supporting
evidence, we observed no user-initiated escalations related to explanation
quality during the online test.

% ===============================================================
\section{Phase IV: Monitoring, Keeping the Judge Aligned}
\label{sec:stewardship}

A judge well-aligned at deployment will not stay aligned. The catalog
shifts (new categories, seasonal content), and the meaning of
``high-quality'' itself can evolve. Phase~IV closes the loop. Each week we
draw a $\sim$300-explanation sample from the generated pool for
HITL evaluation, stratified across the judge's decision
outcomes (\emph{served without revision}, \emph{served after revision},
\emph{dropped}) and biased toward newer catalog items where drift is most
likely. The stratification is fixed week to week, so successive weeks are
comparable. Human raters label the sample under the guidelines of
\S\ref{sec:birth}. The relabeled sample drives drift detection and is
also appended to the train/validation pool, keeping the benchmark live
(the \emph{continuous benchmark augmentation} of
Figure~\ref{fig:lifecycle}), subsampled to preserve class balance. Each
explanation is reviewed by at least three raters and we take the majority
label as ground truth. That panel gives us both a consensus reference and
a direct measure of rater disagreement, and that disagreement is what
sets the bar for the judge.

For each alignment metric
$M$ of \S\ref{sec:metrics}, we score the judge and each individual
rater against the same majority label on the same weekly sample. This
yields one value $M(J)$ for the judge and a set
$M(H)=\{M(h_1),\dots,M(h_R)\}$ of per-rater values, $R\geq3$, whose
mean and standard deviation we compute directly across raters. We
require
\begin{equation}
  M(J) \;\geq\; \mathrm{mean}\big(M(H)\big)
       \;-\; 2\,\mathrm{sd}\big(M(H)\big),
  \label{eq:envelope}
\end{equation}
that is, the judge must score no worse than two standard deviations
below the average rater, where the standard deviation measures
disagreement among the raters themselves. Falling below the band on any
metric raises an \emph{alert}.

Equation~\ref{eq:envelope} is deliberately not a fixed threshold. A harder
weekly sample raises human disagreement, widening $\mathrm{sd}(M(H))$ and
with it the acceptance band, so the judge is not penalized for difficulty
that humans also find hard. We apply the criterion both to the full
weekly sample and separately to newly added titles, where catalog shift
appears first and where a threshold calibrated on established content
would misfire. Requiring no degradation on new titles is what makes the
loop a shift detector rather than a generic regression test.

A drift event triggers Phase~II re-tuning on the augmented benchmark, and
the new rubric is staged behind a manual review gate before deployment,
with the previous rubric retained for rollback. Across every weekly
sample drawn during the large-scale online A/B test (\S\ref{sec:abtest}), the judge
stayed within the human band on every metric, including on newly added
titles, so no re-tuning was triggered. The loop still matters going forward because the catalog,
recommender algorithms, and user population keep changing, so today's
agreement says little about tomorrow's.

Beyond the aggregate metric, the weekly human review repeatedly
surfaced failure patterns that a per-criterion agreement score alone
would not flag. Some are borderline cases whose pass/fail depends on how
confidently the explanation is phrased rather than on the underlying
recommendation quality. Others are concentrated in specific genres, as
with stand-up comparisons where two specials share surface-level tags but
sit in very different cultural contexts. A third group passes every
must-have criterion yet still reads as confusing or oddly framed. None of
these necessarily move the agreement metrics, yet each calls for a change
to the rubrics themselves. When a pattern recurs, our writing and review
experts revise the labeling guidelines of \S\ref{sec:birth}, and we
re-tune the affected judges against the revised guidelines through
Phase~II. Because those guidelines are the source of truth for both the
raters and the judges' seed rubrics, one revision updates both sides of
the comparison at once. Continuous review surfaces the qualitative, long-tail issues
that quantitative drift detection is blind to, catching emerging gaps in
the rubric itself and not just degradation relative to it.

% ===============================================================
\section{Takeaways}
\label{sec:takeaways}

Operating this lifecycle at scale surfaced three lessons we expect to
transfer to other teams shipping LLM judges. First, \emph{invest in the
benchmark before the judge}. A modest, rationale-annotated benchmark is
worth more than a large label-only one, because the rationales are what
make reasoning-aligned tuning possible. Second, \emph{one tuned judge can
serve many roles}. Reusing it for gating and for generation critique
reduces alignment cost and keeps online behavior consistent. Third,
\emph{plan monitoring from day one}. A recommender platform's items and
users constantly evolve, so its LLM judges, like any lifelong agent, must
be monitored and re-tuned.

% ===============================================================
\section{Limitations and Future Work}
\label{sec:future}

\subsection{Limitations}
\label{sec:limitations}

Our evaluation has several scope boundaries. All of it comes from
controlled experiments running over months rather than from continuous
operation, so we cannot say how the lifecycle behaves across the
year-scale shifts it is designed for. The A/B test
(\S\ref{sec:abtest}) covers one surface (mobile) and one explanation
family (similarity-based), so we do not know whether the same lifecycle
and magnitude of lift transfer elsewhere. It also reports only the
metrics we judged most relevant to discovery. RART is validated against a
label-only ablation but not against general-purpose textual optimizers
such as GEPA or TextGrad. The drift-triggered re-tuning path has not yet
fired online, so its automated response is validated only offline.
Appendix~\ref{app:limitations} covers the rest.

\subsection{Future Work}
\label{sec:future-work}

We see four natural extensions of RART. First, recasting rubric tuning
as \emph{memory management}. A benchmark too large to reflect over in one
pass would be consumed in chunks, with the tuner consolidating reasoning
from earlier chunks into a persistent scratchpad that steers each update.
That scratchpad is a \emph{long-term memory}, which new Phase~IV
rationales would update incrementally instead of triggering a full
re-tune, complemented by a \emph{short-term memory} of few-shot examples
retrieved per instance. Second, reasoning-finetuning the judge on
accumulated rationales while text-space rubric optimization handles
online adaptation. Third, casting reasoning-aligned tuning as
\emph{reflective prompt evolution} in the style of
GEPA~\citep{agrawal2026gepa}, of which RART is a greedy special case; a
Pareto pool would trade off our three metrics directly.
Fourth, promoting the judge from evaluator to \emph{reward model}. A
rubric-tuned judge already emits a reason and a label, which is the shape
a reasoning reward model needs, and the same artifacts could post-train
the generator rather than only gate and critique it at inference time.
Here the asymmetry of \S\ref{sec:metrics} suggests a \emph{constrained}
objective rather than the weighted score $s$ (Eq.~\ref{eq:objective}).
Each must-have criterion would be held at its operating point as a
constraint while the softer criteria are optimized, instead of fixing the
trade-off in advance through $w_s$.
% ===============================================================
\bibliography{custom}
\bibliographystyle{colm2026_conference}

% ===============================================================
\appendix

\section{Additional Limitations}
\label{app:limitations}

Beyond the boundaries stated in \S\ref{sec:limitations}, three further
caveats apply. We withhold several system details for
confidentiality, including the specific definitions of the must-have
criteria and the generator and judge model identities. Our meta-judge
validation (\S\ref{sec:reasoning-alignment}) measures agreement on
rationale-agreement classification, a narrower and simpler task than full
explanation judging, so it should not be read as evidence that the
primary judge matches human judgment at a comparable rate. Finally, the
primary judge and the rationale meta-judge $\mathcal{M}$ are built on the
same base model family, so their errors are plausibly correlated; the
human validation of $\mathcal{M}$ bounds this concern but does not remove
it.

\end{document}